\documentclass{article}

    \PassOptionsToPackage{numbers, compress}{natbib}
 \usepackage[dblblindworkshop, final]{neurips_2025}

\workshoptitle{Machine Learning and and The Physical Sciences}

\usepackage[utf8x]{inputenc} % allow utf-8 input
\usepackage[T1]{fontenc}    % use 8-bit T1 fonts
\usepackage{hyperref}       % hyperlinks
\usepackage{url}            % simple URL typesetting
\usepackage{booktabs}       % professional-quality tables
\usepackage{amsfonts}       % blackboard math symbols
\usepackage{nicefrac}       % compact symbols for 1/2, etc.
\usepackage{microtype}      % microtypography
\usepackage{xcolor}         % colors
\usepackage{graphicx}       % for figures
\usepackage{siunitx}        % for units

\title{Data-efficient U-Net for Segmentation of Carbide Microstructures in SEM Images of Steel Alloys}

\author{%
    Alinda Ezgi Gerçek \\
    Helmholtz AI Team Matter (FWCC-A)\\
    Helmholtz-Zentrum Dresden-Rossendorf HZDR\\
    01328 Dresden Germany \\
    \And
    Till Korten \\
    Helmholtz AI Team Matter (FWCC-A)\\
    Helmholtz-Zentrum Dresden-Rossendorf HZDR\\
    01328 Dresden Germany \\
    \And
    Paul Chekhonin \\
    Institute of Resource Ecology, Structural Materials Department\\
    Helmholtz-Zentrum Dresden-Rossendorf HZDR\\
    01328 Dresden Germany \\
    \And
    Maleeha Hassan \\
    Helmholtz AI Team Matter (FWCC-A)\\
    Helmholtz-Zentrum Dresden-Rossendorf HZDR\\
    01328 Dresden Germany \\
    \And
    Peter Steinbach \\
    Helmholtz AI Team Matter (FWCC-A)\\
    Helmholtz-Zentrum Dresden-Rossendorf HZDR\\
    01328 Dresden Germany \\
    \texttt{t.korten@hzdr.de} \\
    }

\begin{document}

\maketitle

\begin{abstract}
    Understanding reactor-pressure-vessel steel microstructure is crucial for predicting mechanical properties, as carbide precipitates both strengthen the alloy and can initiate cracks. In scanning electron microscopy images, gray-value overlap between carbides and matrix makes simple thresholding ineffective.
    We present a data-efficient segmentation pipeline using a lightweight U-Net (30.7~M parameters) trained on just \textbf{10 annotated scanning electron microscopy images}. Despite limited data, our model achieves a \textbf{Dice-Sørensen coefficient of 0.98}, significantly outperforming the state-of-the-art in the field of metallurgy (classical image analysis: 0.85), while reducing annotation effort by one order of magnitude compared to the state-of-the-art data efficient segmentation model.
    This approach enables rapid, automated carbide quantification for alloy design and generalizes to other steel types, demonstrating the potential of data-efficient deep learning in reactor-pressure-vessel steel analysis.
\end{abstract}

\section{Introduction}
\label{sec:introduction}

The mechanical performance of reactor pressure-vessel (RPV) steels -- and ferritic steel components in general -- is influenced by secondary phases within the matrix. Among these, carbide precipitates have a dual effect: they hinder dislocation motion, increasing yield strength, whereas the larger carbides (about \SI{0.5}{\micro\meter} equivalent circle diameter or above)  may act as preferential sites for crack nucleation, promoting brittle failure and reducing fracture toughness \cite{curry_effects_1978,zhang_cleavage_1999,lee_effect_2002,chekhonin_microstructural_2023}. Quantitative descriptors such as carbide number density, size distribution, and spatial arrangement are thus essential for physically-based models of strength and toughness.

Scanning electron microscopy (SEM) of mechanically polished cross-sections remains the standard for visualizing carbides, providing high-resolution, contrast-rich images over large fields of view. In practice, however, gray-level intensities often overlap with surrounding ferritic grains and particle edges may appear blurred. Classical image-analysis pipelines (e.g., denoising, background removal, thresholding and watershed instance segmentation combined with morphological operations) therefore produce often fragmented or spurious segmentation, while manual delineation of each carbide -- though accurate -- requires many hours of expert labor and impractical for the thousands of carbides needed for statistically robust analyses.

Advances in deep convolutional neural networks, especially the encoder-decoder U-Net architecture, have demonstrated state-of-the-art performance on biomedical and materials-science segmentation tasks when large annotated datasets are available \cite{Ronneberger_2015} and even relatively small datasets with hundreds of samples can lead to satisfactory results \cite{bardis_deep_2020}. Unfortunately, generating extensive pixel-wise labels, meaning manually delineated markings of carbides, for SEM micrographs is prohibitively expensive, and there is currently no systematic study investigating how few labeled examples are sufficient to achieve reliable carbide segmentation.

\begin{figure}[htb]
    \centering
    \includegraphics[width=1\textwidth]{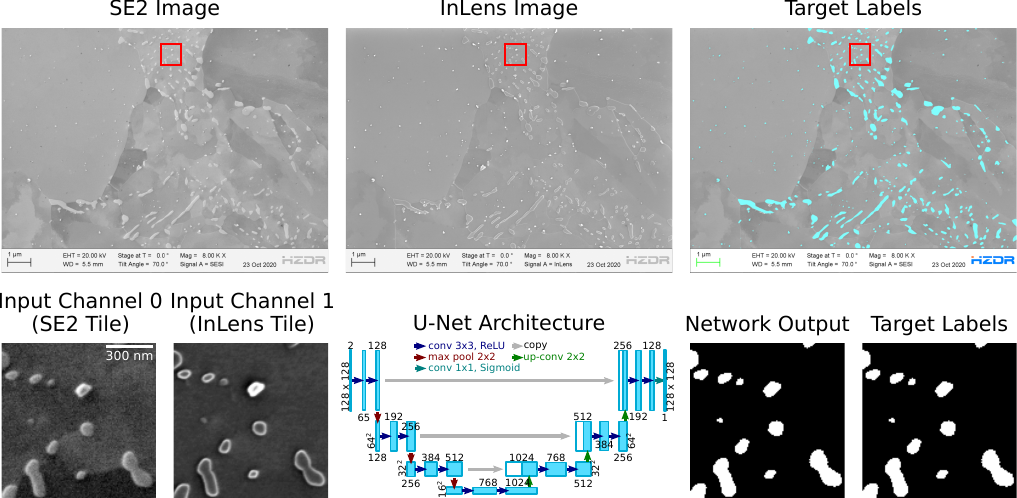}
    \caption{\textbf{Data and model architecture.} SEM images (top) of a RPV steel (JFL) acquired with SE and InLens detectors (top left and top center, respectively) and the corresponding manually annotated target label mask is overlaid in cyan onto the SEM image (top right). Red squares indicate the example tile shown in the bottom row as the two input channels (bottom left) for the U-net (bottom center; adapted from \cite{Ronneberger_2015}) and the network output and corresponding labels (bottom right).}
    \label{fig:1}
\end{figure}

In this work we address the data-efficiency gap by training a lightweight U-Net (30.7~M trainable parameters; see Figure \ref{fig:1} for the network architecture) on only 10  of the 13 available manually annotated SEM images (image size 2048~x~1404~pixels; example in Figure \ref{fig:1}). Our contributions are fourfold: (i) a data-efficient training strategy reduces annotation effort by an order of magnitude, (ii) provide uncertainty estimates by calibrating the output of the network to represent the model's confidence, using temperature scaling, (iii) benchmarking against a handcrafted classical image-analysis baseline, and (iv) demonstrating that the trained network generalizes to SEM images of a different steel (ANP-3).

\section{Methodology}
\label{sec:methodology}

\paragraph{Dataset}
The dataset consists of 13 pairs of SEM images of three RPV steels (material codes: JFL, ANP-10 and ANP-3). Images were acquired using Zeiss NVision 40 (JFL) and Zeiss Ultra 55 (ANP-10) microscopes with two secondary electron detectors (termed SE and InLens from now on) at 2048~x~1404 pixel resolution. The image width corresponds to \SI{14.3}{\micro\meter} and \SI{11.5}{\micro\meter} for JFL and ANP-10, respectively. More details on these steels can be found in \cite{lai_microstructural_2023,chekhonin_microstructural_2023}.

Target label masks were created by merging SE and InLens images (ratio 0.5), applying an initial gray-value threshold, and manually correcting to ensure accurate carbide outlines. The annotation process took approximately 20 hours total.

Preprocessing comprised cropping metadata bars, normalizing pixel intensities to [0, 1], and applying data augmentation (random rotations, flips, Gaussian noise, and blur). 12 images were tiled into 1920 non-overlapping 128~x~128 pixel tiles (randomly split into 80\% training, 10\% validation, 10\% testing). One complete image acquired for a different steel type (ANP-3) was held out to evaluate generalization performance. Code and data to reproduce the results are available at \cite{data, code}.

\paragraph{Classical Baseline}
A classical image-analysis pipeline was implemented using the python wrapper for the image processing library simpleitk \cite{SITK_2013, SITK_2018}. The pipeline merges the two imaging modalities (ratio 0.5), applies Gaussian denoising ($\sigma=1.0$~pixels), performs background removal via top-hat filtering (radius=30~pixels), and segments carbides using Otsu's thresholding method that minimizes the combined variance of foreground and background \cite{Otsu_1979}. Post-processing includes filling holes that are not connected to the boundary and removing small components (< 3 pixels) to eliminate noise.

\paragraph{Network Architecture and Training}
The segmentation model uses a U-Net architecture \cite{Ronneberger_2015} with an encoder-decoder structure and skip connections. Our implementation features a 3-block encoder with 2D convolutions, batch normalization and ReLU activations, connected to a matching decoder via a bottleneck layer. The network has 30.7 Million parameters, making it relatively lightweight and suitable for small datasets. We used a dice loss function:

\begin{equation}
    \mathcal{L}_{Dice} = 1 - \frac{2 \sum_{i=1}^{N} y_i \hat{y}_i}{\sum_{i=1}^{N} y_i + \sum_{i=1}^{N} \hat{y}_i}
\end{equation}

where $y_i$ and $\hat{y}_i$ are target and predicted labels for pixel $i$, respectively.

The network was trained using the Adam optimizer with initial learning rate 0.0002 and learning rate decay (factor 0.5, patience 7 epochs). Early stopping was applied with 14 epochs patience, and training used batch size 32, using $\approx 33$ GB of RAM on one NVIDIA A100 GPU.

\begin{figure}[htb]
    \begin{minipage}[t]{0.68\textwidth}
        \vspace{0pt}
        \includegraphics[width=1.0\textwidth]{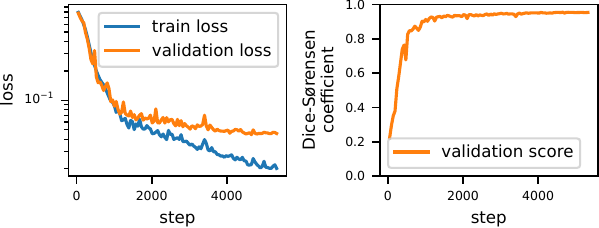}
    \end{minipage}\hfill
    \begin{minipage}[t]{0.3\textwidth}
        \vspace{-5pt}
        \caption{\textbf{Model training.} (left) Training and validation loss curves over 110 epochs. (right) Dice-Sørensen coefficient on the validation set.}
        \label{fig:2}
    \end{minipage}
\end{figure}

Hyperparamter optimization was performed using optuna \cite{akiba2019optuna} over the following parameters: Starting learning rate, early stopping patience, number of features in the first encoder block and number of encoder blocks. The best hyperparameters were selected based on validation set performance (see Appendix \ref{sec:opt}).

In total (including hyperparameter optimization) the training consumed approximately 30 hours on a single NVIDIA A100 GPU; corresponding to a power consumption of approximately 12 kWh.

\paragraph{Uncertainty Estimation}
Particularly for a small dataset -- which may have gaps in distribution coverage -- it is important to be able to assess how certain the model is about it's predictions. This uncertainty information can be useful for downstream tasks, such as active learning or decision-making based on the model's confidence.
To calibrate the model's output probabilities, we applied temperature scaling \cite{guo_calibration_2017} on the validation set. It is a post-hoc method where the single scalar temperature parameter $T$ is learned on a held-out validation set by minimizing the negative log-likelihood, thereby softening the predicted probabilities. The temperature parameter $T$ scales the logits before sigmoid activation:
$\hat{y}_i = \sigma\left(\frac{z_i}{T}\right)$,
where $z_i$ is the logit for pixel $i$. The optimal $T=1.87117$ was determined by minimizing negative log-likelihood on the validation set using L-BFGS \cite{liu_limited_1989}, an optimization algorithm that uses an estimate of the inverse Hessian matrix with a lower memory profile than BFGS. The use of L-BFGS leverages the fact that $T$ is a single scalar and the objective is smooth and deterministic, giving fast convergence with minimal hyperparameter tuning: From a starting value of 1, the algorithm converged to the optimal temperature in just 7 steps. See Appendix \ref{sec:uncertainty} for details.

\paragraph{Statistics}
For statistical hypothesis testing we used the nonparametric Wilcoxon signed-rank test \cite{Wilcoxon_1945} with a significance level of $\alpha = 0.001$ for rejecting the null hypothesis that both samples were drawn from the same distribution.

\paragraph{Code and Data Availability}
Code and data to reproduce the results are available at \cite{data, code}.

\section{Results}
\label{sec:results}

\begin{figure}[htb]
    \centering
    \includegraphics[width=1\textwidth]{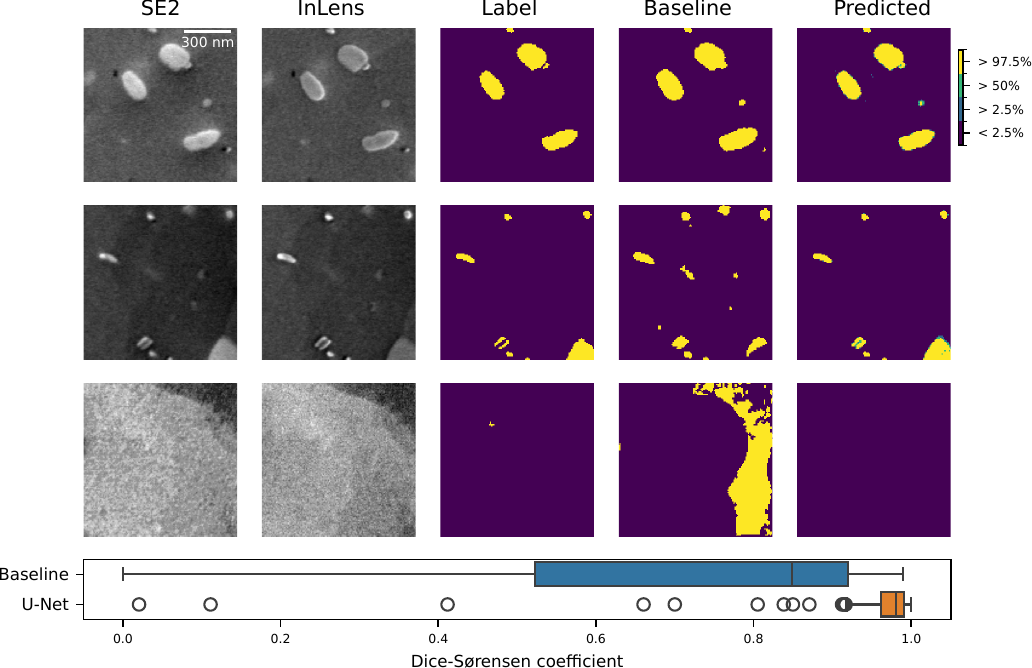}
    \caption{\textbf{Segmentation results.} (top) Example segmentation results on the test set. The model accurately segments carbides of varying sizes and shapes. (bottom) Box plot of Dice-Sørensen coefficients on the test set for the classical image-analysis baseline and our U-Net model. The U-Net significantly outperforms the baseline (Wilcoxon signed-rank test $p<0.001$).}
    \label{fig:3}
\end{figure}

The model was evaluated on a held-out test set of 192 image tiles that were not used during training or validation. Randomly chosen example tiles are shown in Figure \ref{fig:3}. Segmentation performance was quantified using the Dice-Sørensen coefficient \cite{dice_measures_1945,sorensen_method_1948} defined as $\frac{2 TP}{2TP+FP+FN}$ where TP is true positive, FP is false positive and FN is false negative. The U-Net model achieved a median (and interquartile range) Dice-Sørensen coefficient of 0.98 (0.964 - 0.991) on the test set, significantly outperforming the classical image-analysis baseline, which attained a median Dice-Sørensen coefficient of 0.85 (0.522 - 0.919) (Wilcoxon signed-rank test $p<0.001$; see Figure \ref{fig:3}). The model's confidence was lower in areas where it was wrong (green pixels in the predicted confidence maps in Figure \ref{fig:3}), indicating that the uncertainty estimates are meaningful (see also Appendix \ref{sec:uncertainty})).

\begin{figure}[h!tb]
    \centering
    \includegraphics[width=1\textwidth]{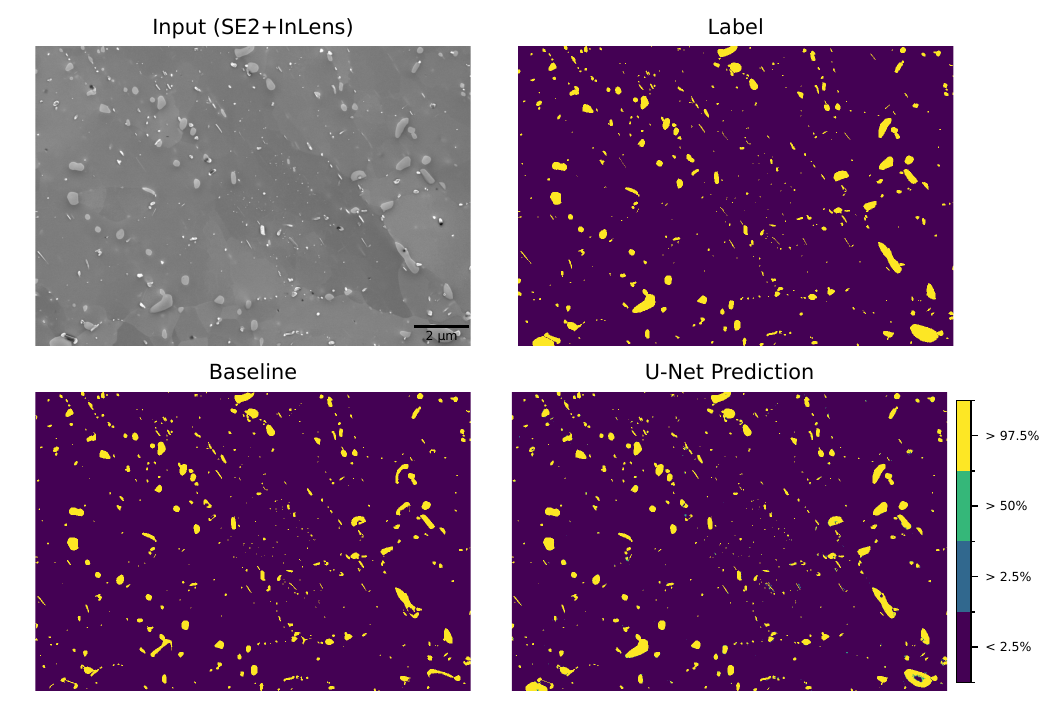}
    \caption{\textbf{Model generalization.} Application of the trained U-Net to SEM images from a different steel (ANP-3). Accurate segmentation of carbides confirms that the model generalizes beyond the training dataset. The baseline achieved a Dice-Sørensen coefficient of 0.90, while the U-Net achieved 0.94.}
    \label{fig:4}
\end{figure}

To test how the model performs on a dataset acquired of a different sample, it was applied to a SEM image of a different steel type (ANP-3) acquired on a separate experiment day (Figure \ref{fig:4}). The U-Net successfully segmented carbides in this image, still outperforming the baseline (Dice-Sørensen coefficient of 0.94, vs 0.90), demonstrating robustness and generalization to previously unseen experimental conditions.

\section{Discussion and Outlook}
\label{sec:discussion}

The results show that high-fidelity segmentation of carbide precipitates in SEM images of RPV steels can be achieved with a lightweight U-Net trained on only 10 annotated images, a 10 fold reduction in labeling effort compared to the state-of-the-art data efficient image segmentation model \cite{bardis_deep_2020}. Extensive data augmentation and a carefully designed training strategy mitigated over-fitting, allowing the model to generalize well to unseen data. The U-Net substantially outperformed the classical image-analysis baseline in both median Dice-Sørensen coefficient and consistency across tiles. Unlike the classical image-analysis baseline, the model does not require manual corrections, demonstrating the model's potential to automate labor-intensive microstructural quantification tasks. The model maintained high accuracy on SEM images from a different steel (ANP-3), highlighting robustness and practical generalization to other steel types.

\paragraph{Limitations}
While the model generalized well to a different imaging session, further validation on larger, more diverse datasets is needed to confirm robustness. Additionally, the current approach relies on fully supervised learning with manually annotated labels, which are time-consuming to generate.

\paragraph{Outlook}
Future work could explore semi-supervised or self-supervised methods to leverage unlabeled data and reduce annotation effort further.

\begin{ack}
We thank the entire Team of Helmholtz AI Matter for invaluable discussions and a great working atmosphere.

\subsection*{Author contributions}
All authors contributed to writing the manuscript. In addition:
\begin{description}
\item[AG and TK]{conducted data processing, trained the machine learning model, and prepared the experimental figures.}
\item[PC] {prepared the samples, acquired the SEM images and created label data}
\item[MH] {performed uncertainty quantification}
%\item{JRS} prepared the FLASH machine and support the project.
\item[PS]{coordinated and provided guidance on machine learning and uncertainty quantification techniques}
\end{description}

\subsection*{Funding}
The work of TK and PS was funded by Helmholtz Incubator Platform Helmholtz AI. The work of TK and MH was funded by the trustPAIKON SAB project.
\end{ack}

\bibliography{references}

\newpage

\appendix

\section{Appendix / supplemental material}
\subsection{Uncertainty Estimation}
\label{sec:uncertainty}

\begin{figure}[htb]
    \centering
    \includegraphics[width=1\textwidth]{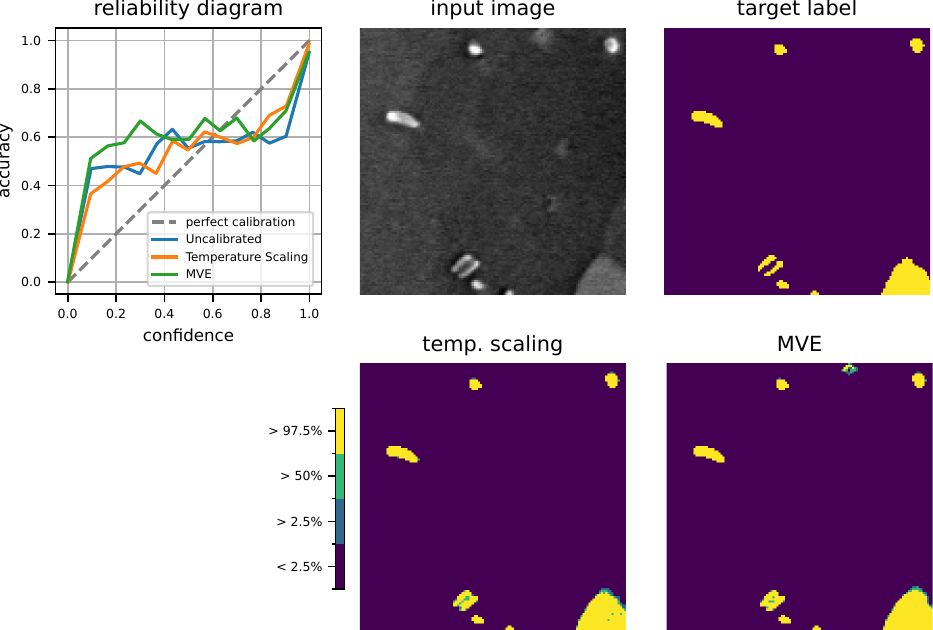}
    \caption{\textbf{Uncertainty estimation} (top right) Reliability diagram before and after temperature scaling. The reliability diagram shows the relationship between predicted probabilities and observed frequencies of the positive class (carbide pixels). The diagonal line represents perfect calibration, where predicted probabilities match observed frequencies. After applying temperature scaling with $T=1.87117$ (orange line), the predictions are better calibrated than without any calibration (blue line) and also better calibrated than with mean-variance estimation (green line), as shown by the curve being closer to the diagonal line. (top center) input image, (top right) target label, (bottom center) model output after temperature scaling, (bottom right) model output with mean-variance estimation. The colors represent the model's confidence in its predictions, with dark green representing pixels classified as true with low confidence, light green medium confidence and yellow high confidence (as indicated in the color bar in the bottom right).}
    \label{fig:S1}
\end{figure}

\paragraph{Temperature scaling}
To estimate the uncertainty of the model's predictions, we applied temperature scaling \cite{guo_calibration_2017} to calibrate the output probabilities. The temperature parameter $T$ was optimized on the validation set by minimizing the negative log-likelihood using the L-BFGS optimization algorithm \cite{liu_limited_1989}. The optimal temperature was found to be $T=1.87117$. The calibration process ensures that the predicted probabilities better reflect the true likelihood of a pixel belonging to the carbide class. Figure \ref{fig:S1} shows the reliability diagram before and after temperature scaling, demonstrating improved calibration of the model's output probabilities (smaller difference to perfect calibration).

\paragraph{Mean-variance estimation}
To estimate the aleatoric (data) uncertainty, we applied mean variance estimation by adding another output layer representing the variance of the predicted logits. Following the approach of \citet{kendall_what_2017}, we used a Monte-Carlo approach to estimate the Dice loss of a set of predictions drawn randomly from a Gaussian distribution with the predicted bias and variance of the respective logit. See the code \cite{code} for implementation details.

The model's confidence was lower in areas where it was wrong (see Figure \ref{fig:S1}), indicating that the uncertainty estimates are meaningful. This uncertainty information can be useful for downstream tasks, such as active learning or decision-making based on the model's confidence. Notably, the reliability diagram (Figure \ref{fig:S1} bottom left) shows that even after calibration, the model is under-confident in a confidence-range of 0.1 -- 0.5 and over-confident in a confidence-range of 0.5 -- 0.9. This is likely caused by high labeling noise around the boundaries of objects. This noise stems from the fact that the labels were generated by a thresholding operation, which is affected by truly random imaging noise and hence inherently unpredictable.

\subsection{Hyperparamter Optimization}
\label{sec:opt}

We used the optuna framework \cite{akiba2019optuna} to perform hyperparameter optimization over the following parameters:
Starting learning rate, early stopping patience, number of features in the first encoder block and number of encoder blocks. The optimization objective was to minimize the Dice-Sørensen coefficient for the validation set. The optimal hyperparameters found were: Starting learning rate = 0.0002, early stopping patience = 14 epochs, number of features in the first encoder block = 128 and number of encoder blocks = 3.

\begin{figure}[htb]
    \centering
    \includegraphics[width=\textwidth]{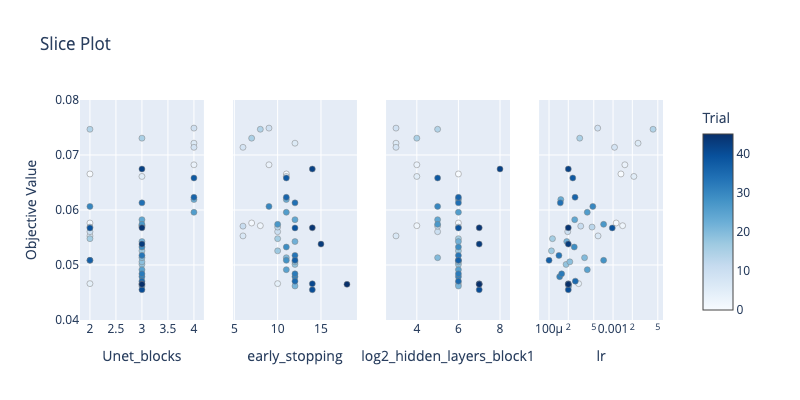}
    \caption{\textbf{Hyperparameter optimization.} Slice plot showing the relationship between hyperparameters and validation loss. Each point represents a trial with a specific combination of hyperparameters, and the color indicates the trial number.}
    \label{fig:S2}
\end{figure}

\end{document}